\newif\ifJOURNAL
\JOURNALfalse
\newif\ifCONF
\CONFfalse
\newif\ifarXiv
\arXivfalse
\newif\ifWP
\WPfalse
\newif\ifFULL
\FULLfalse
\newif\ifLATIN
\LATINfalse

\arXivtrue

\ifarXiv\LATINtrue\fi	
\ifCONF\LATINtrue\fi	

\newif\ifnotCONF	
\notCONFtrue
\ifCONF\notCONFfalse\fi

\newif\ifnotarXiv	
\notarXivtrue
\ifarXiv\notarXivfalse\fi

\newif\ifTR		
\TRfalse
\ifarXiv\TRtrue\fi
\ifWP\TRtrue\fi
\newif\ifnotTR
\notTRtrue
\ifarXiv\notTRfalse\fi
\ifWP\notTRfalse\fi

\newif\ifnotLATIN	
\notLATINtrue
\ifLATIN\notLATINfalse\fi

\ifCONF
  \newcommand{\DFII}{vovk/etal:2005AIStats}
  \newcommand{\DFII}{vovk/etal:2005ALT}
  \newcommand{\DFVIII}{vovk:2006COLT-invited}
\fi
\ifarXiv
  \newcommand{\DFII}{vovk/etal:arXiv0505083}
  \newcommand{\DFIII}{vovk/etal:arXiv0506007}
  \newcommand{\DFVIII}{vovk:arXiv0606093}
\fi
\ifWP
  \newcommand{\DFII}{GTP8}
  \newcommand{\DFIII}{GTP10}
  \newcommand{\DFVIII}{GTP17}
\fi
\ifFULL
  \newcommand{\DFII}{vovk/etal:arXiv0505083}
  \newcommand{\DFIII}{vovk/etal:arXiv0506007}
  \newcommand{\DFVIII}{vovk:arXiv0606093}
\fi

\ifnotLATIN
  \newcommand{\Levin}{levin:1976uniform}
\fi
\ifLATIN
  \newcommand{\Levin}{levin:1976uniform-latin}
\fi

\ifCONF
\documentclass{llncs}
\usepackage{amsmath,amsfonts,amssymb,latexsym,graphicx}
\newcommand{\Extra}[1]{}
\fi

\ifarXiv
\documentclass{article}
\usepackage{amsmath,amsfonts,amssymb,latexsym,graphicx}
\newcommand{\Extra}[1]{}
\fi

\ifWP
\documentclass[toc]{gtarticle}
\usepackage{amsmath,amsfonts,amssymb,latexsym,epsfig,graphicx}
\renewcommand{\Extra}[1]{#1}
\fi

\ifFULL
\documentclass{article}
\usepackage{amsmath,amsfonts,amssymb,latexsym,color,graphicx}
\newcommand{\Extra}[1]{\red{#1}}
\newcommand{\red}[1]{\textcolor{red}{#1}}

\fi

\allowdisplaybreaks[4]

\ifnotLATIN
\input{OT2enc.def}

\fi

\newcommand{\Vladimir}{Vladimir}

\newcommand{\DOT}{.}
\newcommand{\zzrelax}[1]{}

\newcommand{\st}{\mathrel{\!|\!}}

\newcommand{\K}{\mathcal{K}}		
\newcommand{\GGG}{\mathcal{G}}		

\ifnotCONF
\newcommand{\bbbr}{\mathbb{R}}		
\newtheorem{lemma}{Lemma}

\newtheorem{theorem}{Theorem}
\newenvironment{proof}
  {\trivlist\item[\hskip\labelsep\textbf{Proof}]}
  {\endtrivlist}
\fi

\newenvironment{Proof}[1]
  {\trivlist\item[\hskip\labelsep\textbf{Proof #1\;}]}
  {\endtrivlist}
\newcommand{\boxforqed}{\rule{.3em}{1.5ex}}
\newcommand{\qedtext}{\unskip\nobreak\hfil
  \penalty50\hskip1em\null\nobreak\hfil\boxforqed
  \parfillskip=0pt\finalhyphendemerits=0\endgraf}

\newenvironment{remark*}
  {\trivlist\item[\hskip\labelsep{\bfseries Remark}]\relax}
  {\endtrivlist}
\newenvironment{problem*}
  {\trivlist\item[\hskip\labelsep{\bfseries Problem}]\relax}
  {\endtrivlist}

\newlength{\IndentI}
\newlength{\IndentII}
\newlength{\IndentIII}
\newlength{\IndentIV}
\newlength{\WidthI}
\newlength{\WidthII}
\newlength{\WidthIII}
\newlength{\WidthIV}
\title{Defensive forecasting for optimal prediction\\with expert advice}

\ifCONF
\author{Vladimir Vovk}
\institute{Computer Learning Research Centre,
  Department of Computer Science\\
  Royal Holloway, University of London,
  Egham, Surrey TW20 0EX, UK\\
  \email{vovk@cs.rhul.ac.uk}}
\fi

\ifarXiv
\author{Vladimir Vovk\\
\texttt{vovk{\rm@}cs.rhul.ac.uk}\\
\texttt{http://vovk.net}}
\fi

\ifWP
\author{Vladimir Vovk}

\fi

\ifFULL
\author{Vladimir Vovk\\
\texttt{vovk{\rm@}cs.rhul.ac.uk}\\
\texttt{http://vovk.net}}
\fi

\begin{document}
\maketitle
\begin{abstract}
  The method of defensive forecasting
  is applied to the problem of prediction with expert advice
  for binary outcomes.
  It turns out that defensive forecasting
  is not only competitive with the Aggregating Algorithm
  but also handles the case of ``second-guessing'' experts,
  whose advice depends on the learner's prediction;
  this paper assumes that the dependence on the learner's prediction
  is continuous.
\end{abstract}

\section{Introduction}
\label{sec:introduction}

There are many known techniques in competitive on-line prediction,
such as following the perturbed leader
(see, e.g., \cite{hannan:1957,kalai/vempala:2005,hutter/poland:2005}),
Bayes-type aggregation
(see, e.g., \cite{littlestone/warmuth:1994,vovk:1990,freund/schapire:1997a})
and the closely related potential methods,
gradient descent
(see, e.g., \cite{cesabianchi/long/warmuth:1996})
and closely related exponentiated gradient descent
\cite{kivinen/warmuth:1997},
and the recently developed technique of defensive forecasting
(see, e.g., \cite{\DFII,\DFVIII}).
Defensive forecasting combines the ideas of game-theoretic probability
(see, e.g., \cite{shafer/vovk:2001})
with Levin and G\'acs's ideas of neutral measure
\cite{\Levin,gacs:2005}
and Foster and Vohra's ideas of universal calibration \cite{foster/vohra:1998}.
See \cite{cesabianchi/lugosi:2006} for a general review of competitive on-line prediction.

This paper applies the technique of defensive forecasting
to prediction with expert advice in the simple case of binary outcomes
The learner's goal in prediction with expert advice
is to compete with free agents, called experts,
who are allowed to choose any predictions at each step.
We will be interested in performance guarantees of the type
\begin{equation}\label{eq:type}
  L_N
  \le
  \min_{k=1,\ldots,K}
  L_N^k
  +
  a_K
\end{equation}
where $K$ is the number of experts,
$a_K$ is a constant depending on $K$,
$L_N$ is the learner's cumulative loss over the first $N$ steps,
and $L_N^k$ is the $k$th expert's cumulative loss over the first $N$ steps
(see \S\S\ref{sec:quadratic}--\ref{sec:general} for precise definitions).

It has been shown by Watkins
(\cite{vovk:1999derandomizing}, Theorem 8)
that the Aggregating Algorithm
(implementing Bayes-type aggregation for general loss functions
\cite{vovk:1990,vovk:1998game}, the AA for short)
delivers the optimal value of the constant $a_K$ in (\ref{eq:type})
whenever the goal (\ref{eq:type}) can be achieved.
(Watkins's result was based
on earlier results by Haussler, Kivinen, and Warmuth \cite{haussler/etal:1998}, Theorem 3.1,
and Vovk \cite{vovk:1998game}, Theorem 1,
establishing the optimality of the AA
for a large number of experts.)
Theorem \ref{thm:general} of this paper asserts that,
perhaps surprisingly,
defensive forecasting also achieves the same performance guarantee.

Whether the goal (\ref{eq:type}) is achievable
depends on the loss function used for evaluating the learner's and experts' performance.
The necessary and sufficient condition is that the loss function
should ``perfectly mixable''
(see \S\ref{sec:general} for a definition).
For simplicity,
we first consider two specific, perhaps most important,
examples of perfectly mixable loss functions:
the quadratic loss function in \S\ref{sec:quadratic}
and the log loss function in \S\ref{sec:log}.
Those two sections are self-contained
in that they do not require familiarity with the AA.
In the last section, \S\ref{sec:general},
we establish the general result,
for arbitrary perfectly mixable loss functions.
In an appendix
we state Watkins's theorem
in the form needed in this paper.

It is interesting that the technique of defensive forecasting
is also applicable to experts who are allowed to ``second-guess'' the learner:
their recommendations can depend
(in a continuous manner in this paper)
on the learner's prediction.
It is not clear that second-guessing experts can be handled at all
by the AA.

A result similar to this paper's results
is proved by Stoltz and Lugosi in \cite{stoltz/lugosi:2007}, Theorem 14
(a more detailed comparison will be given in \cite{vovk:2007new-local}).
Second-guessing experts are useful in game theory
(where competing with second-guessing experts is known as prediction
with a small internal regret).
For a more down-to-earth example of a useful second-guessing expert,
remember that humans tend to give too categorical (i.e., close to 0 or 1) predictions;
therefore, a useful second-guessing expert for a human learner
would transform his/her predictions to less categorical ones
(according to the learner's expected calibration curve
\cite{dawid:1986}).

\section{Defensive forecasting}
\label{sec:forecasting}

Let $E$ be a topological space
($E=[0,1]^K$ in the application to prediction with expert advice
in \S\S\ref{sec:quadratic}--\ref{sec:general}).

\bigskip

\noindent
\textsc{The binary forecasting protocol}\nopagebreak

\smallskip

\parshape=6
\IndentI  \WidthI
\IndentI  \WidthI
\IndentII \WidthII
\IndentII \WidthII
\IndentII \WidthII
\IndentI  \WidthI
\noindent
$\K_0:=1$.\\
FOR $n=1,2,\dots$:\\
  Expert announces continuous $\gamma_n:[0,1]\to E$.\\
  Forecaster announces $p_n\in[0,1]$.\\
  Reality announces $\omega_n\in\{0,1\}$.\\
END FOR.

\bigskip

\noindent
A \emph{process} is any function $S:(E\times[0,1]\times\{0,1\})^*\to\bbbr$.
Given the sequence of the players' moves in the binary forecasting protocol,
we sometimes write $S_N$, $N\in\{0,1,\ldots\}$, for
$
  S
  \left(
    \gamma_1(p_1),
    p_1,
    \omega_1,
    \ldots,
    \gamma_N(p_N),
    p_N,
    \omega_N
  \right)
$.
(Notice that $S_N$ depend on $\gamma_n$ only via $\gamma_n(p_n)$.)
We also sometimes interpret $S_N$ as function of the players' moves in the protocol
and identify the process $S$ with the sequence of functions $S_N$, $N=0,1,\ldots$,
on the set of all histories $(\gamma_1,p_1,\omega_1,\gamma_2,p_2,\omega_2,\ldots)$.

A process $S$ is said to be a \emph{supermartingale}
if it is always true that
\begin{multline}\label{eq:definition}
  p_N
  S
  \left(
    g_1,
    p_1,
    \omega_1,
    \ldots,
    g_{N-1},
    p_{N-1},
    \omega_{N-1},
    g_N,
    p_N,
    1
  \right)\\
  +
  (1-p_N)
  S
  \left(
    g_1,
    p_1,
    \omega_1,
    \ldots,
    g_{N-1},
    p_{N-1},
    \omega_{N-1},
    g_N,
    p_N,
    0
  \right)\\
  \le
  S
  \left(
    g_1,
    p_1,
    \omega_1,
    \ldots,
    g_{N-1},
    p_{N-1},
    \omega_{N-1}
  \right)
\end{multline}
(i.e., it is true for all $N$,
all $g_1,\ldots,g_N$ in $E$,
all $p_1,\ldots,p_N$ in $[0,1]$,
and all $\omega_1,\ldots,\omega_{N-1}$ in $\{0,1\}$).
In the traditional theory of martingales
(when translated into our framework),
Expert's move is an element of $E$
(in other words, a constant function),
and this would be sufficient for application
to the traditional problem of prediction with expert advice;
however, the version with second-guessing experts requires
the generalization to $\gamma_n:[0,1]\to E$.
We say that a supermartingale $S$ is \emph{forecast-continuous}
if, for each $N$,
$
  S
  \left(
    g_1,
    p_1,
    \omega_1,
    \ldots,
    g_N,
    p_N,
    \omega_N
  \right)
$
is a continuous function of $p_N\in[0,1]$ and $g_N\in E$.
\begin{lemma}[Levin, Takemura]\label{lem:Takemura}
  For any forecast-continuous supermartingale $S$
  there exists a strategy for Forecaster ensuring that $S_0\ge S_1\ge\cdots$
  regardless of the other players' moves.
\end{lemma}
\begin{proof}
  Set, for $p\in[0,1]$ and $\omega\in\{0,1\}$,
  \begin{multline*}
    t(\omega,p)
    :=
    S
    \left(
      \gamma_1(p_1),
      p_1,
      \omega_1,
      \ldots,
      \gamma_{N-1}(p_{N-1}),
      p_{N-1},
      \omega_{N-1},
      \gamma_N(p),
      p,
      \omega
    \right)\\
    -
    S
    \left(
      \gamma_1(p_1),
      p_1,
      \omega_1,
      \ldots,
      \gamma_{N-1}(p_{N-1}),
      p_{N-1},
      \omega_{N-1}
    \right).
  \end{multline*}
  Our goal is to prove the existence of $p$
  such that $t(\omega,p)\le0$ for both $\omega=0$ and $\omega=1$.
  I will give an argument (from \cite{\DFVIII}, the proof of Lemma 1)
  that is applicable very generally.

  For all $p,q\in[0,1]$
  set
  \begin{equation*}
    \phi(q,p)
    :=
    q
    t(1,p)
    +
    (1-q)
    t(0,p).
  \end{equation*}
  The function $\phi(q,p)$ is linear in its first argument, $q$,
  and continuous in its second argument, $p$.
  Ky Fan's minimax theorem
  (see, e.g., \cite{agarwal/etal:2001}, Theorem 11.4)
  shows that there exists $p^*\in[0,1]$ such that
  \begin{equation*}
    \forall q\in[0,1]:
    \quad
    \phi(q,p^*)
    \le
    \sup_{p\in[0,1]}
    \phi(p,p).
  \end{equation*}
  Therefore,
  \begin{equation*}
    \forall q\in[0,1]:
    \quad
    q
    t(1,p^*)
    +
    (1-q)
    t(0,p^*)
    \le
    0,
  \end{equation*}
  and we can see that $t(\omega,p^*)$ never exceeds $0$.
  \qedtext
\end{proof}

For generalizations (due to Levin and Takemura)
of Lemma \ref{lem:Takemura} in different directions,
see, e.g., \cite{\DFIII} (Theorem 1) and \cite{\DFVIII} (Lemma 1).
By defensive forecasting we mean using such results
in prediction with expert advice.

\section{Algorithm competitive with continuous second-guessers:
  quadratic loss function}
\label{sec:quadratic}

This is the version of the standard protocol of prediction with expert advice
under quadratic loss for continuous second-guessing experts:

\bigskip

\noindent
\textsc{Prediction with expert advice under quadratic loss}\nopagebreak

\smallskip

\parshape=9
\IndentI  \WidthI
\IndentI  \WidthI
\IndentI  \WidthI
\IndentII \WidthII
\IndentII \WidthII
\IndentII \WidthII
\IndentII \WidthII
\IndentII \WidthII
\IndentI  \WidthI
\noindent
$L_0:=0$.\\
$L_0^k:=0$, $k=1,\ldots,K$.\\
FOR $n=1,2,\dots$:\\
  Expert $k$ announces continuous $\gamma_n^k:[0,1]\to[0,1]$, $k=1,\ldots,K$.\\
  Learner announces $p_n\in[0,1]$.\\
  Reality announces $\omega_n\in\{0,1\}$.\\
  $L_n:=L_{n-1}+(p_n-\omega)^2$.\\
  $L_n^k:=L_{n-1}^k+(\gamma_n^k(p_n)-\omega_n)^2$.\\
END FOR.

\bigskip

\noindent
To apply Lemma \ref{lem:Takemura} to the problem of prediction with expert advice
under quadratic loss,
we will need the following result.
\begin{lemma}\label{lem:supermartingale-quadratic}
  Suppose $E=[0,1]$ and $\kappa\in[0,2]$.
  The process
  \begin{equation*}
    S_N
    :=
    \exp
    \left(
      \kappa
      \sum_{n=1}^N
      \left(
        \left(
          p_n - \omega_n
        \right)^2
        -
        \left(
          \gamma_n(p_n) - \omega_n
        \right)^2
      \right)
    \right)
  \end{equation*}
  is a supermartingale
  in the binary forecasting protocol.
\end{lemma}
\begin{proof}
  By (\ref{eq:definition}),
  it suffices to check that
  \begin{equation*}
    p
    \exp
    \left(
      \kappa
      \left(
        \left(
          p - 1
        \right)^2
        -
        \left(
          g - 1
        \right)^2
      \right)
    \right)
    +
    (1-p)
    \exp
    \left(
      \kappa
      \left(
        \left(
          p - 0
        \right)^2
        -
        \left(
          g - 0
        \right)^2
      \right)
    \right)
    \le
    1
  \end{equation*}
  for all $p,g\in[0,1]$.
  If we substitute $g=p+x$,
  the last inequality will reduce to
  \begin{equation*}
    p
    e^{2\kappa(1-p)x}
    +
    (1-p)
    e^{-2\kappa px}
    \le
    e^{\kappa x^2},
    \quad
    \forall x\in[-p,1-p].
  \end{equation*}
  The last inequality is a simple corollary of Hoeffding's inequality
  (\cite{hoeffding:1963}, (4.16),
  which is true for any $h\in\bbbr$:
  cf.\ \cite{cesabianchi/lugosi:2006}, Lemma A.1).
  Indeed, applying Hoeffding's inequality to the random variable
  \begin{equation*}
    X
    :=
    \begin{cases}
      1 & \text{with probability $p$}\\
      0 & \text{with probability $1-p$},
    \end{cases}
  \end{equation*}
  we obtain
  \begin{equation*}
    p
    e^{h(1-p)}
    +
    (1-p)
    e^{-hp}
    \le
    e^{h^2/8},
  \end{equation*}
  which the substitution $h:=2\kappa x$ reduces to
  \begin{equation*}
    p
    e^{2\kappa(1-p)x}
    +
    (1-p)
    e^{-2\kappa px}
    \le
    e^{\kappa^2 x^2 / 2}
    \le
    e^{\kappa x^2},
  \end{equation*}
  the last inequality assuming $\kappa\le2$.
  \qedtext
\end{proof}

Lemma \ref{lem:supermartingale-quadratic} immediately implies
a performance guarantee for the method of defensive forecasting.
\begin{theorem}\label{thm:quadratic}
  There exists a strategy for Learner
  in the quadratic-loss protocol with $K$ experts that guarantees
  \begin{equation}\label{eq:quadratic}
    L_N
    \le
    L_N^k
    +
    \frac{\ln K}{2}
  \end{equation}
  for all $N=1,2,\ldots$ and all $k\in\{1,\ldots,K\}$.
\end{theorem}
\begin{proof}
  Consider the binary forecasting protocol with $E=[0,1]^K$.
  By Lemma~\ref{lem:supermartingale-quadratic},
  the process
  \begin{equation*}
    \sum_{k=1}^K
    \exp
    \left(
      \kappa
      \sum_{n=1}^N
      \left(
        \left(
          p_n - \omega_n
        \right)^2
        -
        \left(
          \gamma_n^k(p_n) - \omega_n
        \right)^2
      \right)
    \right)
  \end{equation*}
  is a supermartingale.
  By Lemma~\ref{lem:Takemura},
  Learner has a strategy that prevents this supermartingale from growing.
  This strategy ensures
  \begin{equation*}
    \sum_{k=1}^K
    \exp
    \left(
      \kappa
      \sum_{n=1}^N
      \left(
        \left(
          p_n - \omega_n
        \right)^2
        -
        \left(
          \gamma_n^k(p_n) - \omega_n
        \right)^2
      \right)
    \right)
    \le
    K,
  \end{equation*}
  which implies, for all $k\in\{1,\ldots,K\}$,
  \begin{equation*}
    \exp
    \left(
      \kappa
      \sum_{n=1}^N
      \left(
        \left(
          p_n - \omega_n
        \right)^2
        -
        \left(
          \gamma_n^k(p_n) - \omega_n
        \right)^2
      \right)
    \right)
    \le
    K,
  \end{equation*}
  i.e.,
  (\ref{eq:quadratic}) in the case $\kappa=2$.
  \qedtext
\end{proof}

For the proof of (\ref{eq:quadratic}) being the performance guarantee
for the AA, see, e.g., \cite{vovk:1990}, Example 4, or \cite{vovk:2001competitive}, \S2.4.
It is interesting that even such an apparently minor deviation from the AA
as replacing the AA-type averaging of the experts' predictions
by the arithmetic mean
(with the same exponential weighting scheme)
leads to a suboptimal result:
the constant $2$ in (\ref{eq:quadratic}) is replaced by $1/2$
(\cite{kivinen/warmuth:1999}, reproduced in \cite{vovk:2001competitive}, Remark 3).

\section{Algorithm competitive with continuous second-guessers:
  log loss function}
\label{sec:log}

The log loss function is defined by
\begin{equation*}
  \lambda(\omega,p)
  :=
  \begin{cases}
    -\ln p & \text{if $\omega=1$}\\
    -\ln(1-p) & \text{if $\omega=0$},
  \end{cases}
\end{equation*}
where $\omega\in\{0,1\}$ and $p\in[0,1]$;
notice that the loss function is now allowed to take value $\infty$.
The protocol of prediction with expert advice becomes:

\bigskip

\noindent
\textsc{Prediction with expert advice under log loss}\nopagebreak

\smallskip

\parshape=9
\IndentI  \WidthI
\IndentI  \WidthI
\IndentI  \WidthI
\IndentII \WidthII
\IndentII \WidthII
\IndentII \WidthII
\IndentII \WidthII
\IndentII \WidthII
\IndentI  \WidthI
\noindent
$L_0:=0$.\\
$L_0^k:=0$, $k=1,\ldots,K$.\\
FOR $n=1,2,\dots$:\\
  Expert $k$ announces continuous $\gamma_n^k:[0,1]\to[0,1]$, $k=1,\ldots,K$.\\
  Learner announces $p_n\in[0,1]$.\\
  Reality announces $\omega_n\in\{0,1\}$.\\
  $L_n:=L_{n-1}+\lambda(\omega_n,p_n)$.\\
  $L_n^k:=L_{n-1}^k+\lambda(\omega_n,\gamma_n^k(p_n))$.\\
END FOR.

\bigskip

\noindent
This is the analogue of Lemma \ref{lem:supermartingale-quadratic}
for the log loss function:
\begin{lemma}\label{lem:supermartingale-log}
  Suppose $E=[0,1]$ and $\kappa\in[0,1]$.
  The process
  \begin{equation*}
    S_N
    :=
    \exp
    \left(
      \kappa
      \sum_{n=1}^N
      \Bigl(
        \lambda
        \left(
          \omega_n, p_n
        \right)
        -
        \lambda
        \left(
          \omega_n, \gamma_n(p_n)
        \right)
      \Bigr)
    \right)
  \end{equation*}
  is a supermartingale
  in the binary forecasting protocol.
\end{lemma}
\begin{proof}
  It suffices to check that
  \begin{equation*}
    p
    \exp
    \left(
      \kappa
      \left(
        -\ln p + \ln g
      \right)
    \right)
    +
    (1-p)
    \exp
    \left(
      \kappa
      \left(
        -\ln(1-p) + \ln(1-g)
      \right)
    \right)
    \le
    1,
  \end{equation*}
  i.e., that
  \begin{equation*}
    p^{1-\kappa} g^{\kappa}
    +
    (1-p)^{1-\kappa} (1-g)^{\kappa}
    \le
    1,
  \end{equation*}
  for all $p,g\in[0,1]$.
  The last inequality
  immediately follows from the inequality between the geometric and arithmetic means
  when $\kappa\in[0,1]$.
  (The left-hand side of that inequality is a special case
  of what is known as the Hellinger integral in probability theory.)
  \qedtext
\end{proof}

Lemma \ref{lem:supermartingale-log} implies
a performance guarantee for the log loss function as in the previous section.
\begin{theorem}\label{thm:log}
  There exists a strategy for Learner
  in the log loss protocol with $K$ experts that guarantees
  \begin{equation}\label{eq:log}
    L_N
    \le
    L_N^k
    +
    \ln K
  \end{equation}
  for all $N=1,2,\ldots$ and all $k\in\{1,\ldots,K\}$.
\end{theorem}
\begin{proof}
  Take $\kappa:=1$.
  Lemma~\ref{lem:supermartingale-log}
  guarantees that the process
  \begin{equation*}
    \sum_{k=1}^K
    \exp
    \left(
      \kappa
      \sum_{n=1}^N
      \Bigl(
        \lambda
        \left(
          \omega_n, p_n
        \right)
        -
        \lambda
        \left(
          \omega_n, \gamma_n^k(p_n)
        \right)
      \Bigr)
    \right)
  \end{equation*}
  is a supermartingale in the binary forecasting protocol with $E=[0,1]^K$.
  Any strategy for Learner that prevents this supermartingale from growing ensures
  (\ref{eq:log}) for all $k\in\{1,\ldots,K\}$.
  \qedtext
\end{proof}

For the proof of (\ref{eq:log}) being the performance guarantee
for the AA, see, e.g., \cite{vovk:1990}, Example 3.

\section{Algorithm competitive with continuous second-guessers:
  perfectly mixable loss functions}
\label{sec:general}

In this section we assume that Learner chooses his predictions
from a non-empty \emph{decision space} $\Gamma$
and that his performance is evaluated using a \emph{loss function}
$\lambda:\{0,1\}\times\Gamma\to\bbbr$.
The triple $(\{0,1\},\Gamma,\lambda)$ will sometimes be called our \emph{game of prediction}
(the first element, the outcome space $\{0,1\}$, is redundant at this time).
The loss function will be assumed bounded below;
there is no further loss of generality in assuming that it is non-negative.

As mentioned in \S\ref{sec:introduction},
to have a chance of achieving (\ref{eq:type}),
the loss function has to be assumed ``perfectly mixable''
(this will be further discussed in the appendix);
we start from defining this property.

A point $(x,y)$ of the plane $\bbbr^2$ is called a \emph{superprediction}
(with respect to the loss function $\lambda$)
if there exists a decision $\gamma\in\Gamma$ such that
\begin{equation*}
  \lambda(0,\gamma) \le x
  \quad\&\quad
  \lambda(1,\gamma) \le y.
\end{equation*}
Our next assumption about the game of prediction
will be that the superprediction set is closed.

Let $\eta$ be a positive constant
(the \emph{learning rate} used).
A \emph{shift} of the curve $\{(x,y)\st e^{-\eta x}+e^{-\eta y}=1\}$ in $\bbbr^2$
is the curve $\{(x,y)\st e^{-\eta(x+\alpha)}+e^{-\eta(y+\beta)}=1\}$
for some $\alpha,\beta\in\bbbr$
(i.e., it is a parallel translation of $e^{-\eta x}+e^{-\eta y}=1$
in any direction and by any distance).
The loss function is called \emph{$\eta$-mixable}
if for each point $(a,b)$ on the boundary of the superprediction set
there exists a shift of $e^{-\eta x}+e^{-\eta y}=1$
passing through $(a,b)$
such that the superprediction set lies completely to one side of the shift
(it is clear that in this case the superprediction set
must lie to the Northeast of the shift).
The loss function is \emph{perfectly mixable}
if it is $\eta$ mixable for some $\eta>0$.

Suppose $\lambda$ is $\eta$-mixable, $\eta>0$.
Each decision $\gamma\in\Gamma$ can be represented
by the point $(\lambda(0,\gamma),\lambda(1,\gamma))$
in the superprediction set.
The set of all $(\lambda(0,\gamma),\lambda(1,\gamma))$, $\gamma\in\Gamma$,
will be called the \emph{prediction set};
for typical games this set coincides with the boundary of the superprediction set.
As far as the attainable performance guarantees are concerned
(before we start paying attention to computational issues),
the only interesting part of the game of prediction
is its prediction set;
the game itself can be regarded as an arbitrary coordinate system
in the prediction set.
It will be convenient to introduce another coordinate system
in essentially the same set.

For each $p\in[0,1]$,
let $(a_p,b_p)$ be the point $(x,y)$ in the superprediction set
at which the minimum of $p y+(1-p)x$ is attained.
Since $\lambda$ is $\eta$-mixable,
the point $(a_p,b_p)$ is determined uniquely;
it is clear that the dependence of $(a_p,b_p)$ on $p$ is continuous.

We can now redefine the decision space and the loss function as follows:
the decision space becomes $[0,1]$ and the loss function becomes
\begin{equation*}
  \lambda(0,p):=a_p,
  \quad
  \lambda(1,p):=b_p.
\end{equation*}
The resulting game of prediction
is essentially the same as the original game
(one of the minor differences is that,
if the superprediction set has ``corners'',
a decision $\gamma\in\Gamma$ maybe split
into several decisions $p\in[0,1]$ in the new game,
all leading to the same losses).
In the rest of this section,
let us assume that the game of prediction
has been transformed to this \emph{standard} form.
Notice that the new loss function
is a ``proper scoring rule''
(see, e.g., \cite{dawid:1986}).

The protocol of this section formally coincides
with that of the previous section
(although $\lambda$ ranges over a much wider class of loss functions):

\bigskip

\noindent
\mbox{\textsc{Prediction with expert advice in a standard perfectly mixable game}}\nopagebreak

\smallskip

\parshape=9
\IndentI  \WidthI
\IndentI  \WidthI
\IndentI  \WidthI
\IndentII \WidthII
\IndentII \WidthII
\IndentII \WidthII
\IndentII \WidthII
\IndentII \WidthII
\IndentI  \WidthI
\noindent
$L_0:=0$.\\
$L_0^k:=0$, $k=1,\ldots,K$.\\
FOR $n=1,2,\dots$:\\
  Expert $k$ announces continuous $\gamma_n^k:[0,1]\to[0,1]$, $k=1,\ldots,K$.\\
  Learner announces $p_n\in[0,1]$.\\
  Reality announces $\omega_n\in\{0,1\}$.\\
  $L_n:=L_{n-1}+\lambda(\omega_n,p_n)$.\\
  $L_n^k:=L_{n-1}^k+\lambda(\omega_n,\gamma_n^k(p_n))$.\\
END FOR.

\bigskip

\noindent
Lemmas \ref{lem:supermartingale-quadratic} and \ref{lem:supermartingale-log}
carry over to the perfectly mixable loss functions:
\begin{lemma}\label{lem:supermartingale-general}
  Let $\eta>0$,
  $(\{0,1\},[0,1],\lambda)$ be a standard $\eta$-mixable game of prediction,
  $E=[0,1]$, and $\kappa\in[0,\eta]$.
  The process
  \begin{equation*}
    S_N
    :=
    \exp
    \left(
      \kappa
      \sum_{n=1}^N
      \Bigl(
        \lambda
        \left(
          \omega_n, p_n
        \right)
        -
        \lambda
        \left(
          \omega_n, \gamma_n(p_n)
        \right)
      \Bigr)
    \right)
  \end{equation*}
  is a supermartingale
  in the binary forecasting protocol.
\end{lemma}
\begin{proof}
  It suffices to check that
  \begin{equation*}
    p
    \exp
    \left(
      \kappa
      \left(
        \lambda(1,p)
        -
        \lambda(1,g)
      \right)
    \right)
    +
    (1-p)
    \exp
    \left(
      \kappa
      \left(
        \lambda(0,p)
        -
        \lambda(0,g)
      \right)
    \right)
    \le
    1
  \end{equation*}
  for all $p,g\in[0,1]$.
  As $\lambda$ is $\eta$-mixable, it will also be $\kappa$-mixable;
  we will only be using the latter property.
  Using the notation
  $(a,b):=(a_p,b_p)=(\lambda(0,p),\lambda(1,p))$
  and
  $(a',b'):=(a_g,b_g)=(\lambda(0,g),\lambda(1,g))$,
  we can slightly simplify this inequality:
  \begin{equation}\label{eq:star}
    p
    \exp
    \left(
      \kappa
      \left(
        b - b'
      \right)
    \right)
    +
    (1-p)
    \exp
    \left(
      \kappa
      \left(
        a - a'
      \right)
    \right)
    \le
    1.
  \end{equation}

  It is clear that the superprediction set lies to the Northeast
  of the shift
  $e^{-\kappa(x+\alpha)}+e^{-\kappa(y+\beta)}=1$
  of
  $e^{-\kappa x}+e^{-\kappa y}=1$
  that passes through $(a,b)$,
  \begin{equation}\label{eq:1}
    e^{-\kappa(a+\alpha)}+e^{-\kappa(b+\beta)}=1,
  \end{equation}
  and has the tangent at $(a,b)$ orthogonal to $(1-p,p)$,
  \begin{equation}\label{eq:2}
    \left(
      -\kappa e^{-\kappa(x+\alpha)},
      -\kappa e^{-\kappa(y+\beta)}
    \right)_{x:=a,y:=b}
    \propto
    (1-p,p)
  \end{equation}
  (the expression on the left-hand side is the gradient of
  $e^{-\kappa(x+\alpha)}+e^{-\kappa(y+\beta)}$
  at $(a,b)$).
  We can see from (\ref{eq:1}) and (\ref{eq:2}) that
  \begin{equation*}
    e^{-\kappa(a+\alpha)} = 1-p,
    \quad
    e^{-\kappa(b+\beta)} = p.
  \end{equation*}
  Substituting these values for $p$ and $1-p$ in (\ref{eq:star}),
  we transform (\ref{eq:star}) to
  \begin{equation*}
    \quad
    e^{-\kappa(b'+\beta)}
    +
    e^{-\kappa(a'+\alpha)}
    \le
    1,
  \end{equation*}
  which is true:
  the last inequality just says that $(a',b')$ is Northeast of the shift.
  \qedtext
\end{proof}

\begin{theorem}\label{thm:general}
  Let $\eta>0$
  and consider any standard $\eta$-mixable game of prediction
  $(\{0,1\},[0,1],\lambda)$.
  There exists a strategy for Learner
  in the prediction protocol with $K$ experts that guarantees
  \begin{equation}\label{eq:general}
    L_N
    \le
    L_N^k
    +
    \frac{\ln K}{\eta}
  \end{equation}
  for all $N=1,2,\ldots$ and all $k\in\{1,\ldots,K\}$.
\end{theorem}
\begin{proof}
  Take $\kappa:=\eta$
  and proceed as in the proof of Theorem \ref{thm:log}
  (using Lemma \ref{lem:supermartingale-general}
  instead of Lemma \ref{lem:supermartingale-log}).
  \qedtext
\end{proof}

Inequality (\ref{eq:general}) as the performance guarantee for the AA
is derived in \cite{vovk:1990}, Theorem 1.

\subsection*{Acknowledgments}

This work was partially supported by EPSRC (grant EP/F002998/1),
MRC (grant G0301107),
and the Cyprus Research Promotion Foundation.

\appendix
\section*{Appendix: Watkins's theorem}
\label{sec:Watkins}

Watkins's theorem is stated in \cite{vovk:1999derandomizing} (Theorem 8)
not in sufficient generality:
it presupposes that the loss function is perfectly mixable.
The proof, however, shows that this assumption is irrelevant
(it can be made part of the conclusion),
and the goal of this appendix is to give a self-contained statement
of a suitable version of the theorem.

By a \emph{game of prediction} we now mean a triple $(\Omega,\Gamma,\lambda)$,
where $\Omega$ and $\Gamma$ are sets called the outcome and decision space,
respectively,
and $\lambda:\Omega\times\Gamma\to\overline{\bbbr}$ is called the loss function
($\overline{\bbbr}$ is the extended real line $\bbbr\cup\{-\infty,\infty\}$
with the standard topology,
although the value $-\infty$ will be later disallowed).

Partly following \cite{vovk:1998game},
for each $K=1,2,\ldots$ and each $a>0$
we consider the following perfect-information game $\GGG_K(a)$
(the ``global game'')
between two players, Learner and Environment.

\bigskip

\noindent
\textsc{Global game $\GGG_K(a)$}\nopagebreak

\smallskip

\parshape=9
\IndentI  \WidthI
\IndentI  \WidthI
\IndentI  \WidthI
\IndentII \WidthII
\IndentII \WidthII
\IndentII \WidthII
\IndentII \WidthII
\IndentII \WidthII
\IndentI  \WidthI
\noindent
$L_0 := 0$.\\
$L_0^k := 0$, $k=1,\ldots,K$.\\
FOR $n=1,2,\dots$:\\
  Environment chooses $\gamma_n^k\in\Gamma$, $k=1,\ldots,K$.\\
  Learner chooses $\gamma_n\in\Gamma$.\\
  Environment chooses $\omega_n\in\Omega$.\\
  $L_n := L_{n-1} + \lambda(\omega_n,\gamma_n)$.\\
  $L_n^k := L_{n-1}^k + \lambda(\omega_n,\gamma_n^k)$, $k=1,\ldots,K$.\\
END FOR.

\bigskip

\noindent
Learner wins if, for all $N=1,2,\ldots$ and all $k\in\{1,\ldots,K\}$,
\begin{equation}\label{eq:goal}
  L_N \le L_N^k + a;
\end{equation}
otherwise, Environment wins.

\bigskip

\noindent
It is possible that $L_N=\infty$ or $L_N^k=\infty$ in (\ref{eq:goal});
the interpretation of inequalities involving infinities is natural.

For each $K$
we will be interested in the set of those $a>0$
for which Learner has a winning strategy
in the game $\GGG_K(a)$
(we will denote this by ${\rm L}\smile\GGG_K(a)$).
It is obvious that
\begin{equation*}
  {\rm L}\smile\GGG_K(a)
  \;\&\;
  a'>a
  \Longrightarrow
  {\rm L}\smile\GGG_K(a');
\end{equation*}
therefore, 
for each $K$ there exists a unique \emph{borderline value} $a_K$
such that ${\rm L}\smile\GGG_K(a)$ holds when $a>a_K$
and fails when $a<a_K$.
It is possible that $a_K=\infty$
(but remember that we are only interested in finite values of $a$).

These are our assumptions about the game of prediction
(similar to those in \cite{vovk:1998game}):
\begin{itemize}
\item
  $\Gamma$ is a compact topological space;
\item
  for each $\omega\in\Omega$,
  the function $\gamma\in\Gamma\mapsto\lambda(\omega,\gamma)$
  is continuous;
\item
  there exists $\gamma\in\Gamma$ such that,
  for all $\omega\in\Omega$,
  $\lambda(\omega,\gamma)<\infty$;
\item
  the function $\lambda$ is bounded below.
\end{itemize}

We say that the game of prediction $(\Omega,\Gamma,\lambda)$
is \emph{$\eta$-mixable}, where $\eta>0$, if
\begin{multline}\label{eq:mixture}
  \forall \gamma_1\in\Gamma,\gamma_2\in\Gamma,\alpha\in[0,1] \;
  \exists \delta\in\Gamma \;
  \forall \omega\in\Omega
  \colon\\
  e^{-\eta\lambda(\omega,\delta)}
  \ge
  \alpha
  e^{-\eta\lambda(\omega,\gamma_1)}
  +
  (1-\alpha)
  e^{-\eta\lambda(\omega,\gamma_2)}.
\end{multline}
In the binary case, $\Omega=\{0,1\}$,
this condition says that the image of the superprediction set
under the mapping $(x,y)\mapsto(e^{-\eta x},e^{-\eta y})$
is convex,
and it is easy to see that it is equivalent to the definition
used in \S\ref{sec:general}.

It follows from \cite{hardy/etal:1952}
(Theorem 92, applied to the means $\mathfrak{M}_{\phi}$ with $\phi(x)=e^{-\eta x}$)
that if the prediction game is $\eta$-mixable
it will remain $\eta'$-mixable for any positive $\eta'<\eta$.
(For another proof,
see the end of the proof of Lemma 9 in \cite{vovk:1998game}.)
Let $\eta^*$ be the supremum of the $\eta$
for which the prediction game is $\eta$-mixable
(with $\eta^*:=0$ when the game is not perfectly mixable).
The compactness of $\Gamma$ implies that the prediction game is $\eta^*$-mixable.

\begin{theorem}[Chris Watkins]\label{thm:Watkins}
  For any $K\in\{1,2,\ldots\}$,
  \begin{equation*}
    a_K
    =
    \frac{\ln K}{\eta^*}.
  \end{equation*}
  In particular,
  $a_K<\infty$ if and only if the game is perfectly mixable.
\end{theorem}
It is easy to see that ${\rm L}\smile\GGG_K(a_K)$:
this follows both from general considerations
(cf.\ Lemma 3 in \cite{vovk:1998game})
and from the fact that the AA
and this paper's algorithm based on defensive forecasting
(the latter assuming $\Omega=\{0,1\}$)
win $\GGG_K(a_K)=\GGG_K(\ln K/\eta^*)$.
\begin{Proof}{of Theorem \ref{thm:Watkins}}
  The proof will use Theorem 1 of \cite{vovk:1998game}.
  Without loss of generality
  we can, and will, assume $\lambda>1$
  (add a suitable constant to $\lambda$ if needed);
  therefore, Assumption 4 of \cite{vovk:1998game}
  (the only assumption in \cite{vovk:1998game} not directly made in this paper)
  is satisfied.
  In view of the fact that ${\rm L}\smile\GGG_K(\ln K/\eta^*)$,
  we only need to show that ${\rm L}\smile\GGG_K(a)$ does not hold for $a<\ln K/\eta^*$.
  Fix $a < \ln K/\eta^*$.

  Since the two-fold convex mixture in (\ref{eq:mixture})
  can be replaced by any finite convex mixture
  (apply two-fold mixtures repeatedly),
  the point $(1,1/\eta^*)$ belongs to the separation curve
  (set $\beta:=e^{-\eta^*}$ in the definition of $c(\beta)$)
  whereas the point $(1,a/\ln K)$ is Southwest and outside of the separation curve
  (use Lemmas 8--12 of \cite{vovk:1998game}).
  Therefore,
  E ($=$Environment) has a winning strategy in the game $\GGG(1,a/\ln K)$,
  as defined in \cite{vovk:1998game}.
  It is easy to see from the proof of Theorem 1 in \cite{vovk:1998game}
  that the definition of the game $\GGG$ in \cite{vovk:1998game}
  can be modified, without changing the conclusion about $\GGG(1,a/\ln K)$,
  by replacing the line

  \medskip

  \noindent
  \hspace*{6mm}
  E chooses $n\ge1$ \{size of the pool\}

  \medskip

  \noindent
  in the protocol on p.~153 of \cite{vovk:1998game}
  by

  \medskip

  \noindent
  \hspace*{6mm}
  E chooses $n^*\ge1$ \{lower bound on the size of the pool\}\\
  \hspace*{6mm}
  L chooses $n\ge n^*$ \{size of the pool\}

  \medskip

  \noindent
  (indeed, the proof in \S6 of \cite{vovk:1998game}
  only requires that there should be sufficiently many experts).
  Let $n^*$ be the first move by Environment
  according to her winning strategy.

  Now suppose ${\rm L}\smile\GGG_K(a)$.
  From the fact that there exists Learner's strategy ${\cal L}_1$
  winning $\GGG_K(a)$ we can deduce:
  there exists Learner's strategy ${\cal L}_2$
  winning $\GGG_{K^2}(2a)$
  (we can split the $K^2$ experts into $K$ groups of $K$,
  merge the experts' decisions in every group with ${\cal L}_1$,
  and finally merge the groups' decisions with ${\cal L}_1$);
  there exists Learner's strategy ${\cal L}_3$
  winning $\GGG_{K^3}(3a)$
  (we can split the $K^3$ experts into $K$ groups of $K^2$,
  merge the experts' decisions in every group
  with ${\cal L}_2$,
  and finally merge the groups' decisions with ${\cal L}_1$);
  and so on.
  When the number $K^m$ of experts exceeds $n^*$,
  we obtain a contradiction:
  Learner can guarantee
  \begin{equation*}
    L_N
    \le
    L_N^k
    +
    ma
  \end{equation*}
  for all $N$ and all $K^m$ experts $k$,
  and Environment can guarantee that
  \begin{equation*}
    L_N
    >
    L_N^k
    +
    \frac{a}{\ln K}
    \ln(K^m)
    =
    L_N^k
    +
    ma
  \end{equation*}
  for some $N$ and $k$.
  \qedtext
\end{Proof}

\ifWP
  \DFlastpage
\fi
\end{document}